\documentclass[10pt, a4paper]{article}

\usepackage{lrec-coling2024} 
\usepackage{CJKutf8}
\usepackage[OT2,T1, T2A,OT1]{fontenc}
\usepackage{graphicx}
\usepackage{float}
\usepackage{amsfonts,amssymb,amsmath}
\usepackage{makecell}
\usepackage{booktabs}
\usepackage{arydshln}
\usepackage{multirow}

\title{RU22Fact: Optimizing Evidence for Multilingual Explainable Fact-Checking on Russia-Ukraine Conflict}

\name{ \begin{tabular}{c} Yirong Zeng$^{1}$, Xiao Ding$^{1}$\sthanks{*Corresponding authors}, Yi Zhao$^1$, Xiangyu Li$^2$, \\
    Jie Zhang$^2$, Chao Yao$^2$, Ting Liu$^1$, Bing Qin$^1$ \\
    \end{tabular} }

\address{$^{1}$ Research Center for Social Computing and Information Retrieval, Harbin Institute of Technology \\
         $^2$ Academy of Cyber, China \\
         \{yrzeng,xding,yzhao,tliu,qinb\}@ir.hit.edu.cn \\ 
         \{lixiangyu1101,zhangjie9108,yaochao\}@outlook.com
         }

\abstract{
Fact-checking is the task of verifying the factuality of a given claim by examining the available evidence. High-quality evidence plays a vital role in enhancing fact-checking systems and facilitating the generation of explanations that are understandable to humans. However, the provision of both sufficient and relevant evidence for explainable fact-checking systems poses a challenge. To tackle this challenge, we propose a method based on a Large Language Model to automatically retrieve and summarize evidence from the Web. Furthermore, we construct RU22Fact, a novel multilingual explainable fact-checking dataset on the Russia-Ukraine conflict in 2022 of 16K 
samples, each containing real-world claims, optimized evidence, and referenced explanation. To establish a baseline for our dataset, we also develop an end-to-end explainable fact-checking system to verify claims and generate explanations. Experimental results demonstrate the prospect of optimized evidence in increasing fact-checking performance and also indicate the possibility of further progress in the end-to-end claim verification and explanation generation tasks. 
 \\ \newline \Keywords{fact-checking, evidence, explainability, large language models} }

\begin{document}

\maketitleabstract

\section{Introduction}
\begin{figure}
    \centering
    \includegraphics[width=0.48\textwidth]{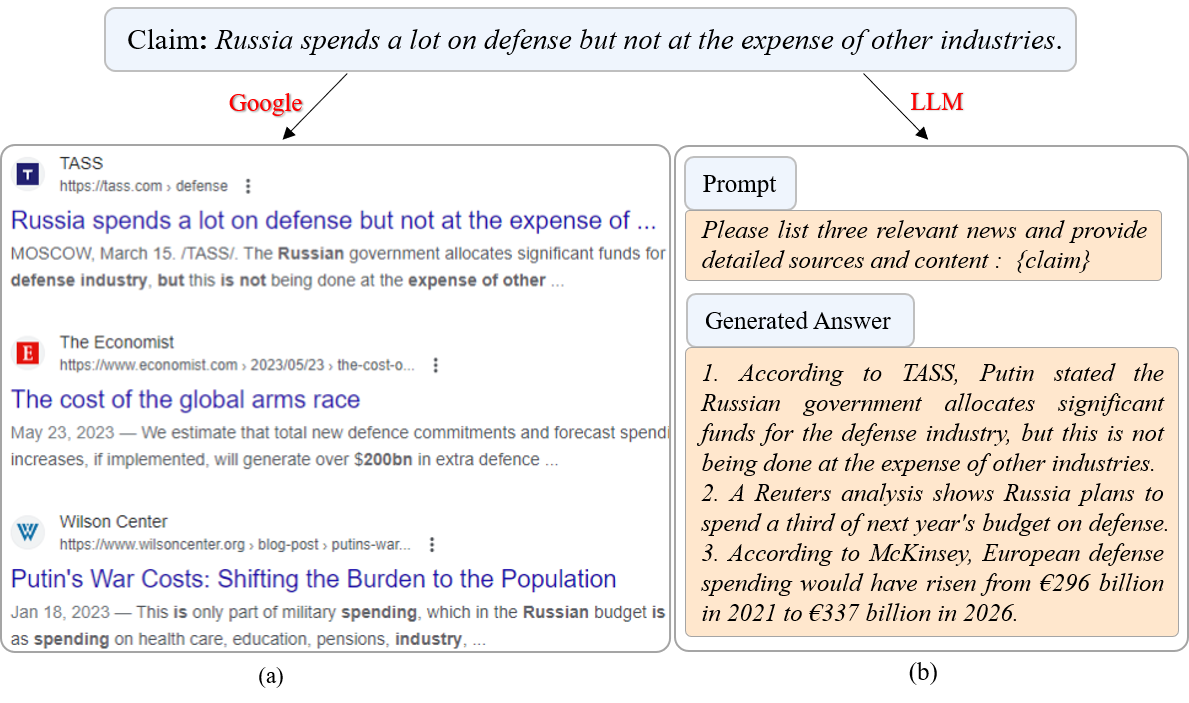}
    \caption{An example of fact verification, based on two different pieces of evidence. (a) a return from a search engine (e.g., Google), and (b) a reply generated by LLMs (e.g., New Bing).}
    \label{fig1}
\end{figure}

As information quickly spreads through social media, fake news become an urgent social issue and even a means of warfare. For example, conspiracy theories about Ukrainian and US bioweapons research during the Russian-Ukrainian conflict emerged \citep{37}.
To combat fake news, automated fact-checking becomes an essential task, which aims to verify the factuality of a given claim based on the collected evidence.
Figure \ref{fig1} (a) illustrates a real-world claim\footnote{\url{https://tass.com/defense/1589173}} that has been verified using a search engine as a basis.

Traditional fact-checking systems follow a pipeline approach that involves an evidence document retrieval module and a claim verification module \citep{14,15}. 
Although most researchers assume that evidence has been properly identified and focus on subsequent steps \citep{6,23,31}, it is crucial to recognize the significant role of evidence in fact-checking \citep{34}.

In fact-checking, it is natural to verify the claim in all the collected documents \citep{38,39}, resulting in a substantial memory footprint due to storage requirements.
To tackle this concern, \citet{17} proposes the extraction of evidence documents from Wikipedia that are relevant to a claim, followed by the selection of the most pertinent sentences from these documents to produce the evidence. 
However, the crowd-sourced claims from this study introduce lexical biases, such as the excessive presence of explicit negation and unrealistic misinformation. 
Recent research retrieves real-world claims from fact-checking websites and considers search snippets \citep{22,36} or retrieved documents \citep{21} provided by search engines as evidence to mitigate this issue. 
Nonetheless, as depicted in Figure \ref{fig1} (a), search snippets often fail to give sufficient information to verify the claim \citep{21}, and the retrieved documents frequently contain a substantial amount of irrelevant information. 
The substandard content retrieved by search engines forces the need for an evidence extractor before the fact verification stage, which would result in error-cascading concerns.
However, providing sufficient and relevant evidence for the fact-checking system is an unresolved challenge. 

In response to the aforementioned challenge, we consider introducing Large Language Models (LLMs) given their excellent performance in natural language understanding \citep{40,41}. As illustrated in Figure \ref{fig1} (b), LLMs have more potential to produce more relevant and sufficient information than search snippets. 
We propose an LLMs-driven method to automatically retrieve and summarize documents from the Web to produce precise evidence with less noise, and refer to the evidence obtained through this method as optimized evidence.
On this basis, we construct RU22Fact, a novel multilingual explainable fact-checking dataset. It contains 16,033 examples, each containing real-world claims, optimized evidence, and referenced explanations about the Russia-Ukraine conflict in 2022. 
We build an explainable fact-checking system to establish the baseline performance and experimental results show that there is room for future improvements in this end-to-end fact-checking and explanation generation task. 
Experimental results demonstrate the prospect of optimized evidence in increasing fact-checking performance, while there is a challenge to solve the problem of generating end-to-end claim verification and explanations\footnote{ Data are available at \url{https://github.com/zeng-yirong/ru22fact}.}.
Our main contributions are summarized as follows: 
\begin{itemize}
    \item We propose an LLMs-driven method to automatically acquire sufficient and relevant evidence from the web. To our knowledge, we are the first to explore optimized evidence in the fact-checking system.

    \item We construct RU22Fact, a novel multilingual explainable fact-checking dataset. This dataset includes optimized evidence to support end-to-end claim verification and human-understandable explanation generation.
\end{itemize}

\section{Related Work}

\subsection{Fact-Checking System}
Normally, when verifying a claim, systems often operate as a pipeline consisting of an evidence document retrieval module, and a claim verification module \citep{14,15}. Most existing methods follow this framework and mainly focus on the last stages \citep{23,6,31}. However, we argue that an optimized evidence document is also critical to building a fact-checking system. At present, there are two main ways to carry out evidence document retrieval. The first is to extract evidence documents related to a claim by entity link \citep{17}, or by TF-IDF \citep{32} from the knowledge base (e.g., Wikipedia) or fact-checking websites, and then select the most relevant sentences from the documents to produce evidence \citep{20,29}. Nevertheless, the source of evidence limits its broad application. The second is to regard search snippets returned by search engines as evidence \citep{22,21}. Although it can verify claims from various sources under real-world scenarios, the low-quality evidence from the search snippet limits the performance of the fact-checking system. Different from these methods, we proposed an automated LLMs-based evidence document retrieval method to produce optimized evidence for building a better fact-checking system. 
\subsection{Fact-Checking Dataset}
We group existing fact-checking datasets into two categories: synthetic and real-world. Synthetic datasets (e.g., Fever \citep{17}, Feverous \citep{18}, Hover \citep{30}), consider Wikipedia as the source of evidence and annotate the sentences of articles as evidence. Although these datasets have made a significant contribution to fact-checking, crowd-sourced claims from this line of work are written with minimal edits to reference sentences, leading to strong lexical biases. Thus, real-world efforts \citep{35,12} extract summaries accompanying fact-checking articles about claims as evidence. Nevertheless, using fact-checking articles restricts evidence to a single source, and they are not available during inference, which is not ideal for developing automated fact-checking systems. To address this issue, some researchers regard search snippets \citep{22,36} or retrieved documents \cite{21} returned by search engines as evidence. However, the low-quality content returned by search engines limits the performance of the system. 
For explainability in the dataset, most existing methods are dedicated to producing extractive explanations (e.g., explanations for veracity predictions about inputs to the system \citep{1,2}), which is unfriendly to humans. 
Recent researchers have formulated the explanation generation task as an abstract summarization problem for human understanding \citep{9,12,13}. 

As shown in Table \ref{tab:relatedworks}, in this paper, we construct a fact-checking dataset, containing real-world claims, high-quality evidence, and referenced explanations, generating explanations as an abstract summarization task.

\begin{table}[]
    \centering
    \small
    \begin{tabular}{lcc}
        \Xhline{1pt}
         \textbf{Dataset} & \textbf{Evi.} & \textbf{Exp.} \\ \Xhline{0.5pt}
         \multicolumn{3}{c}{\textbf{Synthetic}} \\ \hline
         Fever \citep{17}& Wiki. & Ex.\\ 
         Feverous \citep{18} & Wiki. & Ex.\\ 
         Hover \citep{30} & Wiki. & Ex. \\ \hline
         \multicolumn{3}{c}{\textbf{Real-world}} \\ \hline
         MultiFC \citep{36} & FCA & Ex. \\
         PubH \citep{12} & FCA & Ab.\\
         EFact \citep{21} & SE & Ex.\\
         XFact \citep{22} & SE & Ex. \\ \hline
         \textbf{RU22Fact} & LLMs & Ab. \\ \Xhline{0.5pt}
    \end{tabular}
    \caption{Comparison of Fact-Checking Datasets. "Evi." and "Exp." are abbrs for Evidence and Explanation.
    "Ex.", "Ab.", and "Wiki." are abbrs for "Extractive explanation", "Abstract summarization" and "Wikipedia".
    "FCA" and "SE" are abbrs for fact-checking articles and search engines.
    }
    \label{tab:relatedworks}
\end{table}

\section{ Evidence Analysis }
To explore the critical role of evidence in fact-checking, we conducted both manual analysis and experimental analysis. 

In experimental analysis, we conducted an exploratory experiment. XFact \citep{22} is a multilingual fact-checking dataset and contains evidence consisting of documents retrieved by a search engine, which sometimes fails to provide sufficient information for fact-checking. To produce optimized evidence for claims in XFact, we retrieve and summarize documents from the Web for each claim by LLMs. Then we extend XFact with optimized evidence and verify claims based on the evidence from search engines or optimized evidence.
We implement the following experiment according to \citet{22}. 
\begin{enumerate}
    \item \textbf{Attention-based Evidence Aggregator (Attn-EA)}: Aggregation of evidence using an attention-based model that operates on evidence documents retrieved by a search engine. For comparison, we utilize optimized evidence for attention-based evidence aggregator \textbf{ (+OE) }.
    \item  \textbf{Augmenting metadata (+Meta)}: Concatenate additional key-value metadata with the claim text by representing it as a sequence. We also implement an optimized evidence-based evidence aggregator enhanced by metadata \textbf{ (+ Meta + OE)}.
\end{enumerate}

The results are shown in Table \ref{tab:evidence}, from which we can find that the model with optimized evidence achieves better performance compared to the model with retrieved documents by a search engine. This indicates that optimized evidence can provide more sufficient and relevant evidence for fact-checking to improve its performance, and demonstrates the prospect of optimized evidence to solve the fact-checking problem.

\begin{table}[]
    \centering
    {\small
    \begin{tabular}{lccc}
    \Xhline{1pt}
    \textbf{Model}  & ${\alpha_{1} }$ & $ {\alpha_{2} }$ & $ {\alpha_{3} }$  \\ \Xhline{0.5pt}
    Attn-EA     & 38.9 & 15.7 & 16.5 \\
    Attn-EA+Meta & 41.9 & 15.4 & 16.0 \\ \Xhline{0.5pt}
    Attn-EA+OE & 40.37 & \textbf{17.29} & 18.90 \\
    Attn-EA+Meta+OE & \textbf{42.71} & 17.14 & \textbf{19.59} \\ \Xhline{1pt}
    \end{tabular}
    }
    \caption{Average F1 scores of the model. $\alpha_1$ ,$\alpha_2$ and $\alpha_3$ is the different test sets in XFact. $\alpha_1$ is distributionally similar to the training set, $\alpha_2$ is out-of-domain test set and $\alpha_3$ is the zero-shot test set.(\%)}
    \label{tab:evidence}
\end{table}

In manual analysis, we conducted a manual evaluation of the following aspects of evidence: 1) sufficiency: there is sufficient information in the evidence to verify the claim; 2) relevance: each sentence in the evidence relates to the claim. Each aspect is given a score of 1 to 5. 
We compared the original evidence and the optimized evidence in extended XFact and randomly sampled 100 samples for manual evaluation.
We utilize Fleiss' Kappa \citep{fleiss1971measuring} to assess the inter-annotator agreement.
The result is shown in Table \ref{tab:xfact-evidence}, it shows that the optimized evidence obviously outperforms the original evidence in these two aspects, which indicates that the optimized evidence is better at fact-checking.

\begin{table}[]
    \centering
    {\small
    \begin{tabular}{lcc}
    \Xhline{1pt}
    \textbf{Evidence}  & \textbf{Sufficiency} & \textbf{Relevance}  \\ \Xhline{0.5pt}
    Original Evidence     & 2.35(0.53) & 3.02(0.57) \\
    Optimized Evidence & 3.51(0.50) & 4.28(0.63)\\ \Xhline{0.5pt}
    \end{tabular}
    }
    \caption{The manual evaluation of original evidence and optimized evidence. \textbf{Sufficiency} and \textbf{Relevance} represent the average scores of 100 samples. Kappa values are represented in brackets.} 
    \label{tab:xfact-evidence}
\end{table}

\section{Dataset Construction}
In this section, we introduce the whole procedure of dataset construction. 
We construct a multilingual explainable fact-checking dataset, named RU22Fact, of 16k real-world claims related to the 2022 Russia-Ukraine conflict, including conflict coverage, energy crisis, and related stories (e.g., humanitarianism, conspiracy theory, politics). 
To combat fake news about the Russia-Ukraine conflict in different countries and languages, the proposed dataset contains four languages: English, Chinese, Russian, and Ukrainian. An example dataset entry is shown in Table \ref{tab:entry_example}. 

\begin{table}[h]
\centering
{\small
\begin{tabular}{lp{5cm}}
\toprule
\textbf{Claim} & \textit{1,000,000 Ukraine soldiers wiped out.} \\ 
\textbf{Evidence} & \textit{I found a claim on social media that 1,000,000 Ukraine soldiers were wiped out. However, according to a fact-check by PolitiFact, this claim has no official backing. United States and European officials estimate that as many as 120,000 Ukrainian soldiers have died or been injured in the war. } \\ 
\textbf{Explanation} & \textit{United States and European officials estimate as many as 120,000 Ukrainian soldiers have died or been injured in the war. We find no basis for a "1 million" estimate.} \\ \hdashline
Label & Refuted \\
Date &  May 25, 2023 \\
Claimant & Facebook posts \\
Language & English  \\ \bottomrule
\end{tabular}
}
\caption{A example from RU22Fact. Labels and explanations are provided during training but need to be inferred during evaluation.}
\label{tab:entry_example}
\end{table}

\subsection{Data Collection}

To obtain sufficient claims related to the Russia-Ukraine conflict, we collect claims from two sources: fact-checking websites (e.g., Politifact\footnote{ \url{https://www.politifact.com} }, Chinafactcheck\footnote{ \url{https://chinafactcheck.com/}}, Lenta\footnote{ \url{https://lenta.ru/}}) and credible news release websites (e.g., CNN\footnote{\url{https://edition.cnn.com/}}, People's Daily Online\footnote{\url{http://www.people.com.cn/}}, TASS\footnote{\url{https://tass.com/}}). 
We consider websites that included Russian-Ukrainian conflict-related claims and eventually choose ten fact-checking websites and six news release websites, shown in Table \ref{tab:websites}

\begin{table*}[]
    \centering
    {\small
    \begin{tabular}{l|lcc|c}
    \Xhline{1pt}
    \textbf{Source Type}   & \textbf{Website} & \textbf{Quantity } & \textbf{Language} & \textbf{Total}\\ \Xhline{0.5pt}
    \multirow{10}{*}{Fact-checking website} & politifact.com & 2,782 & English &\multirow{10}{*}{6,035} \\ 
                                    & snopes.com & 806 & English &\\
                                    & factcheck.afp.com & 175 & English & \\
                                    & stopfake.org/en & 300 & English &\\
                                    & stopfake.org/uk & 1,082 & Ukrainian & \\
                                    & lenta.ru & 259 & Russian &\\
                                    & factcheck.kz & 58 & Russian &\\
                                    & factpaper.cn & 167 & Chinese &\\
                                    & chinafactcheck.com & 389 & Chinese &\\
                                    & vp.fact.qq.com & 17 & Chinese &\\ 
    \hline
    \multirow{5}{*}{News release website} & edition.cnn.com & 3,890 & English & \multirow{5}{*}{9,998} \\
                                & bbc.com & 739  & English& \\
                                & tass.ru & 2,000  & Russian& \\
                                & pravda.com.ua & 2,430 & Ukrainian \\
                                & people.com.cn & 702  & Chinese& \\
                                & xinhuanet.com & 237  & Chinese& \\
    \Xhline{1pt}
    \end{tabular}
    }
    \caption{The distribution of data sources for the RU22Fact.}
    \label{tab:websites}
\end{table*}
 
 As a starting point, we first query the Russia-Ukraine conflict topic for each website. For websites without such a topic, we search for relevant content using keywords related to the Russia-Ukraine conflict. We scrape fact-checked claims from fact-checking websites and headline claims from news release websites, then take the fact-checking justification from fact-checking websites as referenced explanations for the veracity label of the claim. For headline claims, we summarize the news article by LLMs and check them manually to be referenced explanations. All claims were published between February 2022 and June 2023. In addition to the claim and referenced explanations, we crawl metadata related to each claim such as claimant and date of the claim. Initially, we scraped 39K claims, amounting to 9,037 fact-checked claims from fact-checking websites, and 30,412 news headline claims from news release websites.

\subsection{Data Processing }

\textbf{Dataset Filtering.}\hspace{5pt}
There are two major challenges in using the crawled data directly: 1) standardizing the labels and 2) cleaning the claims and explanations in the dataset. 
The initial data contains 46 labels. Referring to \citet{35}, we review the rating system of the fact-checking websites along with some examples and manually mapped these labels to three categories, including \textit{Supported}, \textit{Refuted}, and \textit{NEI (Not Enough Information)}. For headline claims from news release websites, we assume that they are verified and labeled these \textit{Supported} due to reputable sources, and each claim is assigned one of the three label categories.
In data cleaning, we filter out claims longer than 50 characters to avoid multiple statements in a claim, and we also filter out shorter than 5 words in English, Russian, Ukrainian and 5 characters in Chinese to provide complete semantics in a claim. 
We remove explanations that are less than the length of the claim because it is difficult to provide qualified explanations. 
To alleviate label leakage in some claims, we remove the claims that contain unique keywords associated with the label. 

\vspace{1em} \noindent
\textbf{Optimizing Evidence.}\hspace{1em}
To provide both sufficient and relevant evidence that differs from prior works, we propose an LLMs-driven method to automatically retrieve and summarize documents from the Web to produce optimized evidence. The detailed description is shown in section \ref{sec:OE}.


\subsection{Task Definition}
As we have automatically retrieved and summarized documents to produce optimized evidence, 
We introduce an end-to-end fact-checking approach to verify the claim, instead of a pipeline, taking into consideration the potential issue of error cascading in the pipeline.
Specifically, we explore two subtasks in the proposed dataset, end-to-end claim verification, and explanation generation.
\begin{itemize}
    \item \textbf{Claim Verification}: The Claim Verification task is to predict the label (Supported, Refuted, or NEI) of the claim based on the provided evidence.
    \item \textbf{Explanation Generation}: Given an input claim and optimized evidence, as well as the label, the goal of Explanation Generation is to generate a short paragraph to explain the ruling process and justify the label.
\end{itemize}

 As a result, we collected 16,033 samples covering four languages: English, Chinese, Russian, and Ukrainian. We split the whole dataset into training, development, and test sets. Detailed statistics of the dataset are illustrated in Table \ref{tab:dataset}. Each entry consists of a real-world claim, optimized evidence, referenced explanation, and meta-data (e.g., date, claimant). 
\begin{table}[]
    \centering
    {\small
    \begin{tabular}{lcccc}
    \toprule
    \textbf{Language}  & \textbf{Train} & \textbf{Dev} & \textbf{Test} & \textbf{Total} \\ \midrule
    English   & 6,082 & 867 & 1,741 & 8,690\\
    Chinese   & 1,055  & 152 & 305 & 1,512 \\
    Russia    & 1,621  & 231 & 465 & 3,399 \\
    Ukrainian   & 2,458  & 350 & 704 & 2,430 \\ \hline
    Total     & 11,217  & 1,600  & 3,216 & 16,033\\ \hline \hline
    \multicolumn{2}{l}{\#Support Labels} & \multicolumn{2}{c}{10,081} &\\
    \multicolumn{2}{l}{\#Refuted Labels} & \multicolumn{2}{c}{4,651} &\\ 
    \multicolumn{2}{l}{\#NEI Labels} & \multicolumn{2}{c}{1,301} &\\ \bottomrule
    \end{tabular}
    }
    \caption{Statistics of RU22Fact.}
    \label{tab:dataset}
\end{table}

\section{ Fact-Checking System }
\begin{figure*}
    \centering
    \includegraphics[width=1.0\textwidth,height=0.4\textwidth]{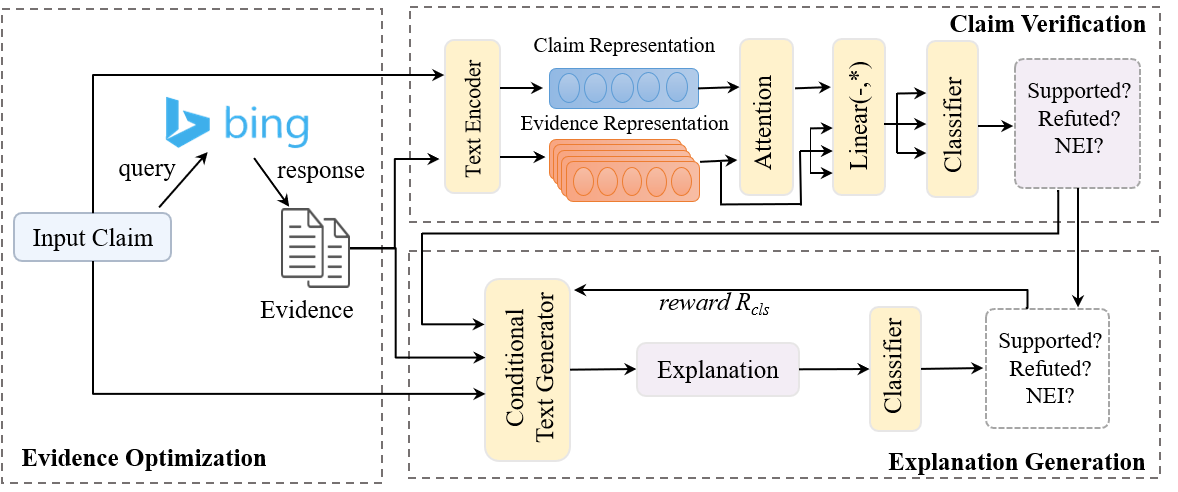}
    \caption{Overview of system framework. It consists of an auto evidence optimization module, a claim verification module, and an explanation generation module.}
    \label{fig:3}
\end{figure*}

In this section, we describe the explainable fact-checking system we built. The framework is illustrated in Figure \ref{fig:3}, which consists of three components: Evidence Optimization, Claim Verification, and Explanation Generation. Next, we will describe the details of each component.

\subsection{Evidence Optimization}
\label{sec:OE}
 We propose an LLMs-driven method to automatically retrieve and summarize retrieved documents from the Web to produce optimized evidence consisting of some sentences. 
 Specifically, it first queries the search engine by a claim and then scratches the retrieved documents. Aiming to make full use of worthy information and remove irrelevant information, subsequently, designing a prompt carefully to summarize the retrieved documents by a single LLM. In practice, we utilize an LLM that can connect to the Internet, such as New Bing\footnote{\url{https://www.bing.com/new}} or Spark\footnote{\url{https://xinghuo.xfyun.cn/spark} }, which can retrieve and summarize documents from the Web, and finish all processes in a single step, taking into account possible error cascades. We query LLM with a carefully designed prompt, such as \textit{"Please list five relevant news and provide detailed sources and content:\{claim\}" }.
 We provide up to five pieces of evidence for each claim in RU22Fact.
 
\subsection{Claim Verification}
Based on the optimized evidence, we further design a claim verification module to predict the truthfulness of each input claim. Given an input claim $C$ and its text evidence $E= \{s_1,s_2,...,s_l \}$, where $s_k$ denotes the $k$-th sentence in evidence, we utilize a text encoder to encode the claim and sentences in the evidence. We feed claim $C$ and sentences $E$ independently to the text encoder and utilize the representations of $CLS$ tokens as their contextual representations: $ X_C \in \mathbb{R} ^ {D} $ and $ X_E = \{ x_{s_1}, x_{s_2},...,x_{s_l}\} \in \mathbb{R} ^ {D \times L} $, where $ D $ denotes the embedding size and $ L $ is the sentence number in the evidence. We then pair each sentence with the input claim and detect the stance of the sentence towards the claim. As Figure \ref{fig:3} shows, we first compute an attention distribution between the claim and the sentence by using $X_C$ as query, $s_k$ as key and value, to compute cross attention and obtain the stance representation $X_{s_k 2C}$. 
$$X_{s_k 2C} = Attention(s_k,X_C).$$

We further obtain the stance representation $H_{s_k 2C}$ of sentence $s_k$ towards claim $X_C$ by concatenating $X_{{s_k 2C}}$ and $s_k$, feeding them to a linear layer:
$$ H_{s_k 2C} = Linear(X_{{s_k 2C}}:s_k),$$
where $[:]$ denotes concatenation operation. In the end, we average the overall stance representation and then feed the result to a linear classifier to predict the label with a cross-entropy objective.

\subsection{Explanation Generation}
To generate a human-understandable explanation for fact-checking prediction, we generate explanations as abstractive summarization and utilize a conditional text generator to generate an explanation by considering the input claim, the predicted label, and the evidence. Further, we incorporate a truthfulness reward based on a classification layer and then optimize the generation model with reinforcement learning to ensure the generated explanation is consistent with the label \citep{13}. As depicted in Figure \ref{fig:3}. Specifically, given an input claim $C$, label $y$, and evidence $E= \{s_1,s_2,...,s_l \}$, we concatenate them into an sequence $X$. Then we feed $X$ as input to conditional text generator and optimize generator for generating explanation $S=\{ s_1,s_2,...s_q\} $ close to the referenced explanation $\widetilde{S} = \{ \widetilde{s_1}, \widetilde{s_2},...,\widetilde{s_q}\}$. We take the gold label as input during training and the predicted label during evaluation. The training objective is to minimize the following negative log-likelihood:
$$ \mathcal{L}_{g}=-\sum_{i} \log (p\left(\tilde{s}_{i} \mid \tilde{s}_{1: i-1}, X ; \phi\right)).$$

To ensure the generated explanation is consistent with the label of the claim, we introduce a truthfulness reward. Specifically, we pre-train a truthfulness classification model, which takes the generated explanation as input and outputs a confidence score for each candidate’s label. 
In practice, we take BERT \citep{42} as a classifier.
$$\boldsymbol{p}(\tilde{y} \mid S)=\operatorname{Softmax}_{i}\left(\operatorname{classifier}_{\theta}(S)\right).$$
We take the difference between the confidence score of the correct answer and the wrong answer as reward $ R_{cls}$ and apply it to policy learning.
$$ R_{c l s}=\boldsymbol{p}\left(\tilde{y}_{C} \mid S\right)-\sum_{\tilde{y}_{j} \neq \tilde{y}_{C} } \boldsymbol{p}\left(\tilde{y}_{j} \mid S\right), $$
where $ \widetilde{y}_{C}$ is the gold label of $C$, $ \widetilde{y}$ and $ \widetilde{y}_{j}$ is the predicted label.

\section{ Experiment }
\label{sec:experiments}
We conduct experiments to evaluate the performance of two tasks: Claim Verification and Explanation Generation in the proposed dataset RU22Fact.
\subsection{Claim Verification}
We adopt three different text encoders: 1) Multilingual BERT ($FCS_{mBert}$), a multilingual variant of BERT \citep{42}. 2) XLM-RoBERTa ($FCS_{XLM-R}$), a multilingual version of RoBERTa pretrained on CommonCrawl data containing 100 languages \citep{XLM-R}. 3) DistilBERT ($FCS_{dBert}$), a distilled version of the BERT based multilingual model \citep{Sanh2019DistilBERTAD}.

To analyze the proposed dataset, we adopt the following different settings to conduct claim verification experiments in the fact-checking system, $FCS$: 1) only consider the claim without evidence, $FCS_{claim}$; 2) only consider evidence without claim, $FCS_{evidence}$; 3) consider the claim with random evidence, $FCS_{re}$. Random evidence denotes random sampling evidence for each claim in RU22Fact.
We utilize $FCS_{mBert}$ as the text encoder for these settings.

The experimental result is assessed against precision (Pr), recall (Rc), and macro F1 metrics. 
The result is shown in Table \ref{tab:claim_verify}. 
The performance of $FCS_{claim}$ performs worse than $FCS_{mBert}$, indicating that the claim lacks sufficient information for claim verification. $FCS_{re}$ performs worse than $FCS_{mBert}$ and similarly to $FCS_{claim}$, indicating there is no obvious bias in the evidence.
When using random evidence, the model tends to focus on the claim rather than the evidence.
$FCS_{evidence}$ performs better than $FCS_{claim}$ and similarly to $FCS_{mBert}$, indicating there is more useful information in the optimized evidence than in the claim.

According to the results, we find that $FCS_{mBert}$ achieves similar performance compared to $FCS_{dBert}$ and $FCS_{XLM-R}$, with a macro F1 score of approximately 60\%. 
However, there is still room for further improvement in the proposed dataset RU22Fact.
The challenges include low resource language processing, and the label distribution is uneven.

\begin{table}[!htbp]
    \centering
    {\small
    \begin{tabular}{lccc}
        \Xhline{1pt}
        \textbf{Settings}  &  \textbf{Pr} & \textbf{Rc} & \textbf{F1} \\ \hline
        $FCS_{claim}$ & 65.71 & 59.01 & 57.93 \\
        $FCS_{evidence}$ & 68.07 & 62.33 & 59.68  \\
        $FCS_{re}$ & 55.33 & 59.84 & 57.35\\ \hline \hline
        $FCS_{XLM-R}$ & 57.56 & \textbf{63.57} & 59.91\\
        $FCS_{dBert}$ & \textbf{74.30} & 63.49 & 60.40\\
        $FCS_{mBert}$ & 58.31 & {62.91} & \textbf{60.56} \\ \hline
        \Xhline{1pt}
    \end{tabular}
    }
    \caption{Performance of Claim Verification. The claim and evidence are concatenated and input into the text encoder in $FCS$, $FCS_{XLM-R}$, $FCS_{dBert}$, $FCS_{mBert}$ represent three different text encoders used in the fact-checking system ($FCS$). (\%) }
    \label{tab:claim_verify}
\end{table}

\subsection{Explanation Generation}
We adopt two different conditional text generators in this section:
1) Bart-large-cnn ($FCS_{bart}$) \citep{DBLP:journals/corr/abs-1910-13461}, a transformer encoder-decoder model with a bidirectional encoder and an autoregressive decoder, fine-tuned on CNN Daily Mail \citep{cnn}. 
2) T5-base ($FCS_{t5}$), a Text-to-Text Transfer Transformer model, which is a versatile and efficient pre-trained model for various natural language processing tasks \citep{2020t5}.
We also add GPT-3.5-turbo-0613 (${GPT3.5}$, an AI chat mode based on the GPT-3.5-series model that generates responses based on user input \citep{openai2022}, as the interpreter generator for comparison.
We fine-tune $FCS_{bart}$ and $FCS_{t5}$ to generate the explanation in the fact-checking system, and prompt ${GPT3.5}$ to generate the explanation.
We use two methods for evaluating the quality of explanations generated: automated evaluation and qualitative evaluation.
\begin{table}[]
    \centering
    {\small
    \begin{tabular}{lcccc}
    \Xhline{1pt}
     \textbf{Settings} & \textbf{Rouge1} & \textbf{Rouge2} & \textbf{RougeL} & \textbf{BLEU} \\ \Xhline{0.5pt}
      $FCS_{bart}$ & 34.17 & 16.28 & 32.08 & 9.56 \\ 
      $FCS_{t5}$ & 32.24 & 15.17 & 30.47 & 8.92  \\
      ${GPT3.5}$ & 36.56 & 18.90 & 34.10 & 17.05  \\
      \Xhline{1pt}
    \end{tabular}
    }
    \caption{ROUGE and BLEU scores for generated explanation via our explainable fact-checking system. (\%)}
    \label{tab:eval}
\end{table}
\subsubsection{ Automated Evaluation }
We evaluate the generated explanation by ROUGE \citep{45} and BLEU \citep{46}, and use the F1 values for ROUGE-1, ROUGE-2, and ROUGE-L.

The results are shown in Table \ref{tab:eval}. 
From the results, we find that fine-tuned $FCS_{bart}$ performs better than $FCS_{t5}$ in the fact-checking system, and ${GPT3.5}$ achieves the best performance because part of the referenced explanation in the dataset comes from itself.
\subsubsection{ Qualitative Evaluation }
Evaluation using ROUGE and BLEU does not present a complete picture of the quality
of these explanations, therefore, we introduce three desirable coherence properties for machine learning explanations and evaluate the quality of the generated explanations against them \citep{12}. 
More about these three coherence properties is shown in Appendix A.

\begin{itemize}
    \item \textbf{Strong Global Coherence}. It holds for a generated fact-checking explanation, every sentence in the explanatory text must entail the claim.
    \item \textbf{Weak Global Coherence}. It holds for a generated fact-checking explanation, no sentence in the explanatory text should contradict the claim (by
entailing its negation).
    \item \textbf{Local Coherence}. The generated explanation satisfies local consistency if each sentence in the explanatory text does not contradict each other. 
\end{itemize}

We employ human evaluation to assess the quality of the referenced explanation and the generated explanations for these properties. We randomly sampled 100
samples from the test set of RU22Fact, and five annotators to evaluate them according to these three properties.

Also, we conduct a computational evaluation of the three properties using NLI (natural language inference \citep{dagan2022recognizing}). In practice, we use two pretrained NLI models: BERT trained in MNLI (the multi-genre natural language inference corpus \citep{williams-etal-2018-broad}) and RoBERTa trained in MNLI.

The results of the qualitative evaluation are shown in Table \ref{tab:compute-eval}. The referenced explanation achieves almost the best results on three properties and three NLI models, which implies that the referenced explanation is of higher quality than others.
RoBERTa trained in MNLI performs better than BERT trained in MNLI, which means RoBERTa is a better approximation of these three properties.
NLI models are reliable approximations of weak global coherence and local coherence, and they seem to be a poor approximation for strong global coherence.
\begin{table}[]
    \centering
    {\small
    \begin{tabular}{lccc}
    \Xhline{1pt}
     \multirow{3}{*}{\textbf{Method} }  & \multirow{2}{*}{\textbf{SGC} } & \multirow{2}{*}{\textbf{WGC} } & \multirow{2}{*}{\textbf{LC} } \\ \\ \cline{2-4}
                                        & \multicolumn{3}{c}{ \textbf{Human} } \\ \Xhline{0.5pt}
      ${Referenced\ Explanation}$ & \textbf{58.23} & \textbf{96.63} & \textbf{93.38}  \\ 
      $FCS_{bart}$ & 43.97 & 92.08 & 90.14  \\ 
      $FCS_{t5}$ & 40.29 & 91.62 & 90.47  \\ 
      ${GPT3.5}$ & 53.29 & 95.73 & 90.52  \\ \hline
      \multicolumn{4}{c}{\textbf{BERT;MNLI} } \\ \hline
      ${Referenced\ Explanation}$ & 37.72 & \textbf{60.17} & \textbf{60.36}  \\ 
      $FCS_{bart}$ & 45.52 & 59.94 & 53.02  \\
      $FCS_{t5}$ & \textbf{46.64} & 58.77 & 49.79  \\ 
      ${GPT3.5}$ & 37.18 & 59.74 & 54.84  \\ \hline
      \multicolumn{4}{c}{ \textbf{RoBERTa;MNLI} } \\ \hline
      ${Referenced\ Explanation}$ & \textbf{24.13} & \textbf{94.96} & \textbf{90.03}  \\
      $FCS_{bart}$ & 20.26 & 93.31 & 87.51  \\
      $FCS_{t5}$ & 19.46 & 94.26 & 85.41  \\ 
      ${GPT3.5}$ & 21.78 & 94.65 & 88.56  \\ \hline
      \Xhline{1pt}
    \end{tabular}
    }
    \caption{The results of the qualitative evaluation in three properties, strong global coherence (\textit{SGC}), weak global coherence (\textit{WGC}), and local coherence (\textit{LC}) properties. (\%)}
    \label{tab:compute-eval}
\end{table}

\section{Conclusion}
In this paper, we first analyze the challenge of providing sufficient and relevant evidence for fact-checking, and then propose an LLMs-driven method to automatically retrieve and summarize documents from the Web to produce optimized evidence. An analytical experiment indicates that optimized evidence can provide more sufficient and relevant information for building a better fact-checking system. Furthermore, we construct a novel multilingual explainable fact-checking dataset named RU22Fact, including real-world claims, optimized evidence, and referenced explanation. To establish the baseline performance, we build an explainable fact-checking system based on RU22Fact. Experimental results demonstrate the prospect of optimized evidence to increase fact-checking performance including claim verification and explanation generation.
\begin{figure}[]
    \centering
    \includegraphics[scale = 0.55]{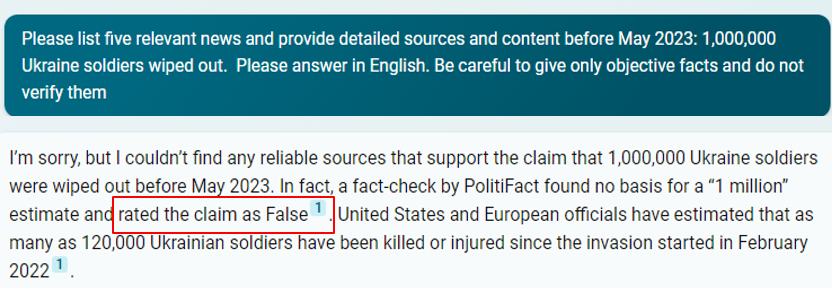}
    \caption{An example of information leakage, and the red box indicates label leakage.}
    \label{fig:info_leakage}
\end{figure}
\section{Limitations}
Several limitations should be considered in this paper, though this paper provides a step forward in fact-checking. 
\begin{itemize}
    \item \textbf{Information Leakage}: There is possible information leakage when retrieving documents from the web. 
    To alleviate this problem, we add some restrictions to a prompt, such as "\textit{Be careful to give only objective facts and do not verify them}". 
    Nevertheless, It sometimes fetches snippets of fact-checking articles if the claim comes from a fact-checking website, which can also lead to information leakage. 
    An example is shown in Figure \ref{fig:info_leakage}, there is a label leakage in the red box. 


    \item \textbf{Low-resource Languages}: The dataset proposed in this work covers claims related to the Russia-Ukraine conflict of 2022, a worldwide topic that is not limited to high-resource languages. However, our work covers only four languages and has less data in non-English languages, which limits fact-checking in low-resource languages.

    \item \textbf{Domain Generalization}: In this article, our data set is a topic related to the Russia-Ukraine conflict. It might not work well in the same way for other topics, and it requires further research.
    
\end{itemize}

\section{Acknowledgements}
We thank the anonymous reviewers for their constructive comments, and gratefully acknowledge the support of the National Natural Science Foundation of China under Grants U22B2059 and 62176079, Natural Science Foundation of Heilongjiang Province under Grant Y02022F005.

\section{Bibliographical References}\label{sec:reference}

\bibliographystyle{lrec-coling2024-natbib}

\bibliography{lrec-coling2024-example}

\bibliographystylelanguageresource{lrec-coling2024-natbib}
\bibliographylanguageresource{languageresource}

\section*{Appendix}

\subsection*{Appendix A. Coherence Property} \label{sec:annex_a}

\begin{figure*}
    \centering
    \includegraphics[scale=0.9]{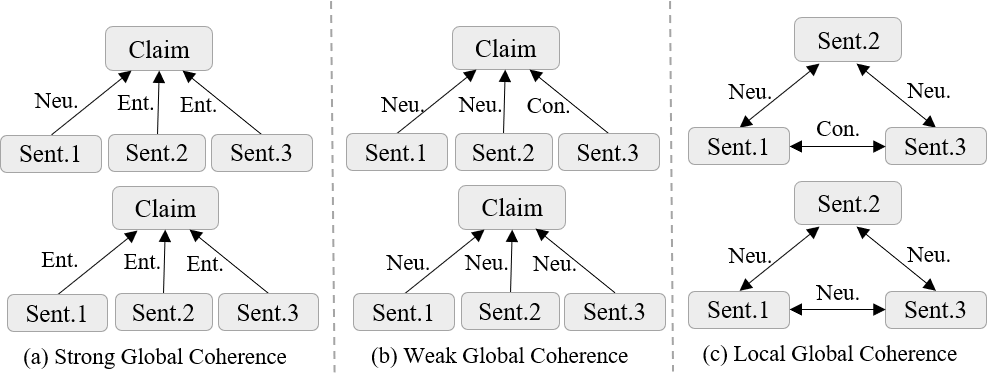}
    \caption{Schematic representations of strong global coherence, weak global coherence and local coherence. 
    "Neu.", "Ent." and "Con." are abbrs for neutral, entails and contradicts.
    In each column, the upper part means coherence cannot be satisfied, and the lower part means coherence is satisfied.}
    \label{fig:swc_three}
\end{figure*}
\begin{figure*}
    \centering
    \includegraphics[scale=0.85]{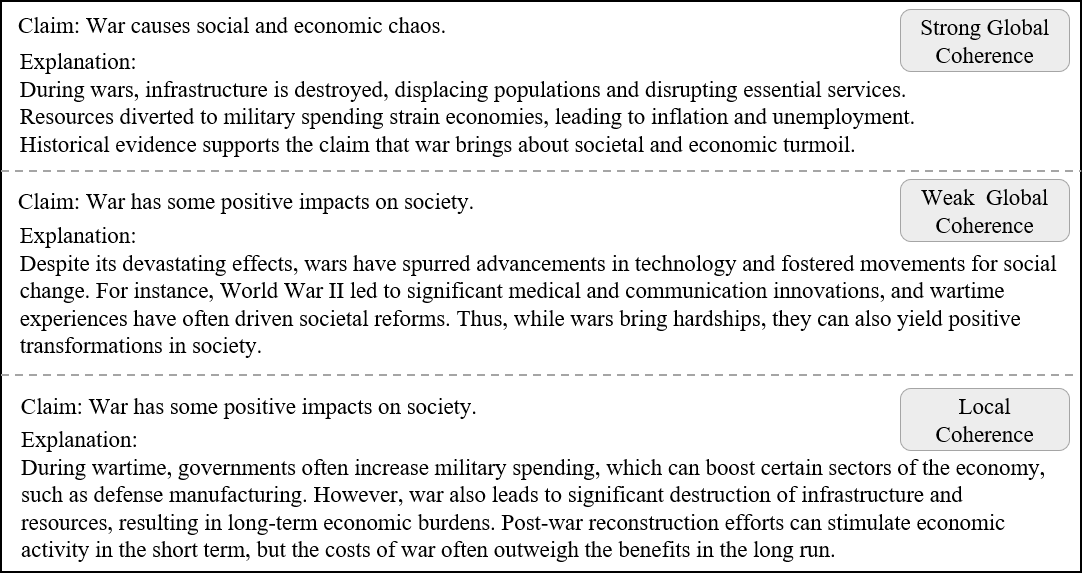}
    \caption{Examples of strong global coherence, weak global coherence and local coherence. }
    \label{fig:coherence}
\end{figure*}

Figure \ref{fig:swc_three} demonstrates the three coherence properties schematically in graphical form.
Figure \ref{fig:coherence} demonstrates examples of the three coherence properties.

\end{document}